# OCR Context-Sensitive Error Correction Based on Google Web 1T 5-Gram Data Set

**Youssef Bassil**
*LACSC – Lebanese Association for Computational Sciences*
*Registered under No. 957, 2011, Beirut, Lebanon*
E-mail: youssef.bassil@lacsc.org

**Mohammad Alwani**
*LACSC – Lebanese Association for Computational Sciences*
*Registered under No. 957, 2011, Beirut, Lebanon*
E-mail: mohammad.alwani@lacsc.org

## Abstract

Since the dawn of the computing era, information has been represented digitally so that it can be processed by electronic computers. Paper books and documents were abundant and widely being published at that time; and hence, there was a need to convert them into digital format. OCR, short for Optical Character Recognition was conceived to translate paper-based books into digital e-books. Regrettably, OCR systems are still erroneous and inaccurate as they produce misspellings in the recognized text, especially when the source document is of low printing quality. This paper proposes a post-processing OCR context-sensitive error correction method for detecting and correcting non-word and real-word OCR errors. The cornerstone of this proposed approach is the use of Google Web 1T 5-gram data set as a dictionary of words to spell-check OCR text. The Google data set incorporates a very large vocabulary and word statistics entirely reaped from the Internet, making it a reliable source to perform dictionary-based error correction. The core of the proposed solution is a combination of three algorithms: The error detection, candidate spellings generator, and error correction algorithms, which all exploit information extracted from Google Web 1T 5-gram data set. Experiments conducted on scanned images written in different languages showed a substantial improvement in the OCR error correction rate. As future developments, the proposed algorithm is to be parallelised so as to support parallel and distributed computing architectures.

## 1. Introduction

Approximately, more than 200 million paper books are being published every year (Vincent, 2007). However, since a computer can only process digital data, paper-based texts must be converted into digital format. Generally, e-books short for electronic books are electronic version of a printed book (Oxford Dictionary, 2010), consisting of both text and images, and are readable by computers. It was estimated that over 2 million e-books are available for download on the Internet (World eBook Fair, 2009). E-books require less storage space than paper books, they can also be replicated many times, shared online, and digitally processed easily, mainly searched, translated, edited, and annotated (Pastore, 2008). Fundamentally, creating e-books involves translating traditional paper books into electronic files; this process is more known as OCR short for Optical Character Recognition which converts scanned images of text into editable digital electronic documents or e-books (Cheriet, Kharma, Liu, and Suen, 2007). Alas, OCR systems are not that perfect as they are erroneous and exhibit spelling errors in the recognized output text, especially when the images being scanned are of poor printing quality and contain several defects (Hartley and Crumpton, 1999). One way to solve this situation is to manually review and correct the OCR output by hand. Although this approach sounds doable, it is costly, time consuming, laborious, and error-prone as the human eye may miss some mistakes. A more effective approach is to automatically post spell-check the output of the OCR system using a spell-checker coupled with a dictionary of words. The drawback of this approach is that it requires a wide-ranging dictionary comprising a large vocabulary of massive terms, proper names, special jargons and expressions, domain-specific terms, and technical terminologies, enough to cover every single word in the language.

This paper proposes a post-processing OCR error correction method based on web-scale information extracted from Google Web 1T 5-gram data set (Google Inc., 2006). Essentially, the Google 5-gram data set houses a big volume of *n*-gram word statistics up to 5 grams extracted from the World Wide Web and suitable for performing text spelling correction. Integrally, the proposed solution is a blend of three modules: An error detector that detects non-word errors in the OCR text using unigram counts from Google Web 1T data set; a candidate spellings generator based on a character 2-gram model that generates correction suggestions for every single detected error; and a context-sensitive error corrector that corrects OCR errors using 5-gram statistics from Google Web 1T data set.

All in all, the proposed solution is expected to considerably improve the OCR error correction rate and reduce the number of misspellings in the OCR recognized output text.

## 2. Optical Character Recognition

Optical Character Recognition (OCR) is the process of transforming images of handwritten or typewritten text into machine-editable text (Cheriet, Kharma, Liu, and Suen, 2007). For this purpose, a sequence of steps or stages must be executed.

The first stage is the image acquisition, in which paper-based documents are scanned using a modern scanner device connected to a computer system. Often, the scanning activity is done in gray scale mode to ease the subsequent digital processing of the document and the recognition of characters.

The second stage is the pre-processing, in which the scanned image is processed and modified using state-of-the-art digital image processing algorithms such as noise reduction, data normalization, and data compression. In their book (Gonzalez and Woods, 2007), the authors extensively discuss the various methods and algorithms involved in the pre-processing of digital images. As its name implies, noise reduction eliminates all artifacts and imperfections from the scanned image. Besides, filtering techniques such as contrast equalization (Gonzalez and Woods, 2007), sharpening (Leu, 2007), smoothing (Legault and Suen, 1997), and thresholding (Mo and Mathews, 1998) are usually all employed to remove and correct bumps, unclear regions, and blurry areas in the image. Normalization is the process of making text consistent in a way to remove textual deviations in writings, and produce a standardized text (Guerfaii and Plamondon, 1993). One of the techniques engaged in normalization is resizing, stretching, and rotating text in order to make it homogenous. Another technique is data compression whose sole goal is to eliminate the number of white spaces and blank lines between characters and paragraphs; and thus reducing the image size and increasing the OCR processing speed. The outcome of this proactive stage is a polished and flawless image that is easier and faster to manipulate by the OCR system.

The third stage is the segmentation stage, in which the document is divided into smaller components, called feature objects. There exist two types of segmentation: external segmentation which divides a document into blocks of paragraphs, sentences, and words; and internal segmentation which divides a document into individual characters, letters, and numbers, called glyphs (basic blocks representing one or more characters) (Casey and Lecolinet, 1996). Segmentation is regarded as a fundamental procedure in OCR systems because it is the foundation of character recognition as it helps pulling out the typewritten characters from the scanned image so that they can be correctly identified in later stages.

The fourth stage is the feature extraction stage, in which the scanned document is decomposed into several features to facilitate the recognition of characters and words (Trier, Jain, and Taxt, 1996). Characteristically, every character in the alphabet has some unique features that make it exclusively distinguishable among other characters. Therefore, features with similar characteristics belong to the same class and consequently represent a particular distinctive character. The algorithms behind the feature extraction process rely on the topological and geometrical structures of the characters of the alphabet such as width, height, shape, length of strokes, change in curvature, and the relative distance between two distinct points of the same character.

The fifth stage is the classification of features, in which all previously extracted features are classified into independent classes and categories, for instance, all features with similar characteristics belonging to group A are recognized as character A; all features with similar characteristics belonging to group N are recognized as character N. To cope with this formula, a classifier is utilized to map every group of similar features into a real character of the alphabet. The process of classifying features can be done either manually, using straight pattern matching algorithms, or automatically, using an artificial neural network. In the latter approach, a neural network is trained with a set of training data that is actually composed of pre-defined sample features with their corresponding correct classes; several epochs are executed until the network is successfully trained. At run-time, the trained network is used to automatically predict the proper class for the extracted features so as to identify the correct characters and words. Ultimately, the output of this stage is an electronic machine-readable ASCII text that can be handled and managed using a computer machine.

The sixth and last stage is the post-processing, in which the recognized OCR text is examined for spelling and linguistic errors. For this purpose, a lookup dictionary or lexicon is used to spell-check, detect, and correct syntactic mistakes and errors in the recognized output text. Additionally, some OCR systems employ context-based error correction to semantically and grammatically correct OCR errors according to their context in the sentence.

## 3. OCR Post-Processing Error Correction

As mentioned earlier, post-processing is the last stage of an OCR system whose goal is to detect and correct spelling errors in the OCR output text. In effect, two types of spelling errors exist: non-word error which is a misspelling that results in an invalid word; and real-word error which is a misspelling that results in a valid word, however, grammatically or semantically incorrect with respect to its context (Kukich, 1992).

In point of fact, several non-trivial post-processing error correction algorithms exist, one of which is the string matching algorithm that generates candidate spellings for every detected error in the text and then weight them using a distance metric representing various costs. The correction candidate with the lowest distance with respect to the misspelled word is the best to fit as a correction (Levenshtein, 1966).

Guyon and Pereira, (1995) demonstrated that using the language syntactic properties and the *n*-gram model can speed-up the process of generating correction candidates and ultimately picking up the best matching one.

Niwa, Kayashima, and Shimeki (1992) proposed an OCR post error correction method based on pattern learning, wherein a list of correction candidates is first generated from a lexicon. Then, the most proper candidate is selected as a correction based on the vocabulary and grammar characteristics surrounding the error word.

Liu, Babad, Sun, and Chan (1991) proposed a statistical method for auto-correction of OCR errors. The approach uses a dictionary to generate a list of correction candidates based on the *n*-gram model. Then, all words in the OCR text are grouped into a frequency matrix that identifies the exiting sequence of characters and their count. The correction candidate having the highest count in the frequency matrix is then selected to substitute the error word.

Taghva, Borsack, and Condit (1994) proposed an improved design that employs a clustering technique to build a set of groups containing all correction candidates. Then, several iterations of word frequency analysis are executed on these clusters to eliminate the unlikely candidate words. In due course, only a single candidate will survive to replace the misspelled word.

Wick, Ross, and Learned-Miller (2007) proposed the use of a topic model to correct the OCR output text. It is a global word probability model, in which documents are labeled with a semantic topic having a specific independent vocabulary distribution. In other words, every scanned document is semantically classified according to its topic using an unsupervised training model. Every misspelled word is then corrected by selecting the correction candidate that belongs to the same class of the actual error.

Kise, Shiraishi, Takamatsu, and Fukunaga (1996) proposed a divergent approach based on syntactic and semantic correction of OCR errors. The idea pivots around the analysis of sentences to deduce whether or not they are syntactically and semantically correct. If a suspicious sentence is encountered, possible correction candidates are generated from a dictionary and grouped top-down with respect to their strongest syntactic and semantic constraints. In the long run, the candidate on the top of each group is the one that must substitute the corresponding OCR error.

Hull (1996) proposed the idea of using a Hidden Markov Model (HMM) to integrate syntactic information into the post-processing error correction. The suggested model achieved a higher rate of error correction due to its statistical nature in selecting the most probable candidate for a particular misspelled word.

Sun, Liu, Zhang, and Comfort (1992) introduced an intelligent autonomic model able of self-learning, self-configuring, and self-adapting. The idea behind it is that as the system operates, as its ability to self-find and self-correct errors increases.

Borovikov, Zavorin, and Turner (2004) proposed a blend of post-processing tools that assist in the detection and correction of spelling errors. In this method, the OCR text is sent through a series of filters with the intention of correcting misspellings via multiple passes. On every pass, a spell-checker tool intervenes to detect and correct misspelled words. After several passes, the number of OCR errors starts by exponentially getting reduced.

## 4. Proposed Solution

This paper proposes a new post-processing OCR error correction method for detecting and correcting non-word and real-word errors in OCR text based on Google Web 1T 5-gram data set (Google Inc., 2006). The Google 5-gram data set, originally published by Google Inc. at the Linguistic Data Consortium (LDC) contains a colossal volume of data statistics represented as word *n*-gram sequences with their respective frequencies, all extracted from online public web pages. As these data are mined from the Internet, they cover large vocabulary that includes generic words, proper names, domain-specific terms, technical expressions, jargons, acronyms, and terminologies. As a result, it virtually emulates a universal, comprehensive, and multi-domain lexicon, suitable for performing extensive dictionary-based error correction.

Primarily, the proposed OCR error correction method is a mixture of three modules: The first module consists of an OCR error detector that detects non-word errors in the OCR text using Google Web 1T unigram data set (a subset of Google Web 1T 5-gram data set). The second module consists of a candidate spellings generator that generates correction suggestions for every detected error in the OCR text using Google Web 1T unigram data set and a character-based 2-gram model. The third module is an OCR contextual error corrector that selects the best spelling candidate as a spelling correction using 5-gram counts pulled out from Google Web 1T 5-gram data set. Figure 1 depicts the three different modules of the proposed method along with their inner workings.

**Figure 1:** The Proposed Post-Processing OCR Error Correction Method

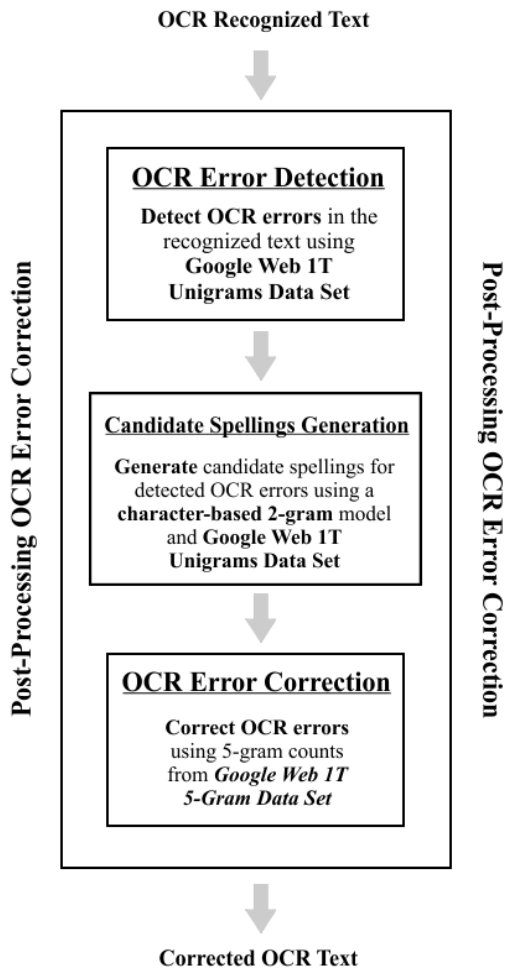

## 4.1. The Error Detection Algorithm

The proposed OCR error detection algorithm detects non-word errors $N=\{ne_1,ne_2,ne_3,ne_p\}$ in the OCR text $O=\{o_1,o_2,o_3,o_q\}$ where *ne* denotes an non-word error word, *p* denotes the total number of detected errors, *o* denotes a word in the OCR text, and *q* denotes the total number of words in the OCR text. The algorithm starts by validating every word $o_i$ in *O* against Google Web 1T unigram data set; if an entry for $o_i$ is found, then $o_i$ is said to be correct and no correction is to take place. In contrast, if no entry for the word $o_i$ is found, then $o_i$ is said to be misspelled; and hence, a correction is required. Eventually a list of errors is built and is represented as $N=\{ne_1,ne_2,ne_3,ne_p\}$ where *p* is the total number of non-word errors detected in the OCR text. Below is the pseudo-code for the proposed OCR error detection algorithm.

    **Function** ErrorDetection (O)
    {
        T ← Tokenize(O," ") *// splits the OCR text on whitespace and returns tokens into array T*

        **while** (i < N) *// validates all tokens*
        {
            *// searches for every T[i] in Google Web 1T unigrams data set*

```
            results ← Search(GoogleDataSet , T[i])

        if(results = true)  // means that an entry for T[i] is found in Google data set → spelled correctly
            i ← i+1  // move to the next token T[i+1]
        else  // means that T[i] is misspelled → requires correction
            SpawnCandidates(T[i])  // executes the candidate spellings generation algorithm
    }
}
```

### 4.2. The Candidate Spellings Generation Algorithm

The proposed candidate spellings generation algorithm constructs a list of probable spelling corrections for the detected non-word errors in the OCR output text. These possible corrections are called candidate spellings and they are denoted as S=$\{s_{11}, s_{12}, s_{13}, s_{1p}, \ldots, s_{u1}, s_{u2}, s_{u3}, s_{uq}\}$ where $s$ denotes a given candidate spelling, $u$ denotes the total number of detected non-word OCR errors, and $p$ and $q$ denote the total number of candidates generated for a given error. At heart, the algorithm is based on a character-based 2-gram model which spawns candidate spellings by looking for unigrams in Google Web 1T data set that share 2-gram character sequences with the error word.

For instance, given that "case where only one sangle element is allowed to be stored" is a particular sentence in the OCR text that needs to be spell-checked. Obviously, the misspelling is "sangle" which is supposed to be "single". Tokenizing "sangle" into 2-gram character sequences would yield to: "sa" , "an" , "ng" , "gl" , "le". The ultimate task is to find a set of unigrams from Google Web 1T data set that contain one or more of these 2-gram sequences. Table 1 shows a possible scenario for the error word "sangle".

**Table 1:** Google Unigrams Sharing 2-Gram Characters with the Error Word "Sangle"

| 2-Gram Sequences | Unigrams from Google Data Set |
|---|---|
| sa | *sa*lute *sa*ndbox *sa*nd *sa*le *sa*ndwich *sa*lt *sa*nitary |
| an | t*an*gle s*an*itary s*an*dbox s*an*d s*an*dwich m*an* *an*gle |
| ng | ta*ng*le si*ng*le E*ng*lish a*ng*le ti*ng*le fri*ng*e ri*ng* |
| gl | sin*gl*e sin*gl*y tin*gl*e an*gl*e bea*gl*e tan*gl*e En*gl*ish |
| le | sing*le* ang*le* beag*le* unab*le* ting*le* tang*le* disab*le* |

Since the results might be hundreds of unigrams, it is more feasible to filter them down and select only the top ten unigrams having the highest number of common 2-gram character sequences with the error word "sangle". Table 2 outlines the list of the top ten candidate spellings.

**Table 2:** List of Top Ten Candidate Spellings

| Candidate Spellings | Number of Shared 2-Gram Character Sequences |
|---|---|
| tangle | 4 |
| angle | 4 |
| single | 3 |
| tingle | 3 |
| beagle | 2 |
| sand | 2 |
| sandbox | 2 |
| English | 2 |
| sanitary | 2 |
| sandwich | 2 |

The complete pseudo-code for the proposed candidate spellings generation algorithm is given below:

```
Function SpawnCandidates (error)
{
    A ← Create2Grams(error)  // create 2-gram character sequences and saves them in array A

    while (i < N)  // processes all 2-gram sequences in array A
    {
        // finds unigrams having A[i] as substring, that is unigrams sharing 2-gram sequence
        // with the error word
        list[i] ← Substring(GoogleDataSet , A[i])

        i ← i+1
    }

    // picks the top ten unigrams sharing 2-gram character sequences with the error word, puts the results in
    // array candidates
    candidatesList ← TOP(list)

    ErrorCorrection(candidatesList)  // executes the error correction algorithm
}
```

### 4.3. The Error Correction Algorithm

The proposed OCR error correction algorithm starts by first taking each spawned candidate spelling $s_{im}$ with 4 words that directly precede the initial error in the OCR output text. Formally, it can be denoted as $L_m=$" $o_{i-4}$ $o_{i-3}$ $o_{i-2}$ $o_{i-1}$ $s_{im}$ " where L represents a sentence made out of 5 words, $o$ denotes a word preceding the initial OCR error, $s$ denotes a particular candidate spelling for a particular error, $i$ denotes the $i_{th}$ word that precedes the initial OCR error, and $m$ denotes the $m_{th}$ candidate spelling. Afterwards, the algorithm tries to find every $L_m$ in the Google Web 1T 5-gram data set. The candidate $s_{im}$ in sentence $L_m$ with the highest count is chosen to substitute the originally detected OCR error word. The proposed algorithm is context-sensitive as it depends on real-world statistics from Google data set, primarily extracted from the World Wide Web. As a result, and back to the previous example, regardless of the fact that the candidate "tingle" is a valid correction for the error word "sangle", candidate "single" would be selected as a correction instead of "tingle" because the sentence "case where only one tingle" has a lower frequency on the Internet than the sentence "case where only one single". Table 3 shows the different $L_m$ 5-gram sentences from Google Web 1T 5-gram data set, each related to a particular candidate spelling.

**Table 3:** 5-Gram Sentences from Google Data Set

| m | $L_m$ 5-Gram Sentences |
|---|---|
| 1 | case where only one tangle |
| 2 | case where only one angle |
| 3 | case where only one single |
| 4 | case where only one tingle |
| 5 | case where only one beagle |
| 6 | case where only one sand |
| 7 | case where only one sandbox |
| 8 | case where only one English |
| 9 | case where only one sanitary |
| 10 | case where only one sandwich |

The complete pseudo-code for the proposed OCR error correction algorithm is given below:

```
Function ErrorCorrection (candidatesList)
{
    while (i < N)  // processes all candidate spellings
    {
        // puts together the ith candidate with the four preceding words
        L ← Concatenate(T[j-4] , T[j-3] , T[j-2] , T[j-1] , candidatesList[i] )

        // finds L in Google 5-gram data set and returns its frequency
        freq[i] ← Search(GoogleDataSet , L)

        i ← i+1
    }

    k ← MaxFreq(freq)  // returns the index k of the candidate whose L has the highest frequency

    RETURN candidatesList[k]  // returns the correction for the OCR error
}
```

## 5. Experiments & Results

Experiments were conducted over two low-quality image documents written in two different languages. OmniPage 17 by Nuance Communications (Nuance Communications, Inc, 2011) was used to perform OCR. It is worth noting that OmniPage also features a built-in post-processor to spell-check the recognized output text after being recognized. For evaluation purposes, the obtained text was respell-checked using the proposed method, and both results were compared head-to-head. The first document is an extract from McFarland (2006) written in the English language; while the second is an extract from Sillamy (2003) written in the French language.

The scanned image of the English document is shown in Figure 2. Table 4 outlines the results of the OmniPage program along with the generated OCR errors which are underlined. Table 5 shows the final results after applying the proposed OCR error correction method.

**Figure 2:**   Low-Quality English Image Document

In 1968, Bob Noyce and Gordon Moore left Fairchild Semiconductor to start their own company focused on building products from integrated circuits. They named their company *Intel* (from INTegrated ELectronics). In 1969, Intel began shipping the first commercial integrated circuit using MOSFETs, a 256-bit memory chip called the 1101. The 1101 memory chip did not sell well, but Intel was able to rapidly shrink the size of the new silicon gate MOSFETs and add more transistors to their designs. One year later Intel offered the 1103 with 1024 bits of memory, and this rapidly became a standard component in the computers of the day.
   In April 1970, Intel hired Faggin, the inventor of the silicon gate MOSFET, away from Fairchild. On Faggin's second day at Intel, Masatoshi Shima, the engineering representative from Busicom, arrived from Japan to review the design.

**Table 4:** Results of Performing OCR using OmniPage

|  |
|---|
| In 1968, Bob **Noycc** and Gordon **Noore leit** Fairchild Semiconductor **ro** start their own **compary** focused on **builnop** products from integrated **cieeuts**. They named their company **Lntelo** (from **INTgrated** Electronics). In 1969, **Tntel** began s**hioping** the first commercial integrated circuit using **MOSRETs**, a 256-bit **mimory ehip** called the **L101**. The 1101 memory chip did not **rell** well, but Intel was able to rapidly shrink the size of the **rew sileon** gate MOSFETs and add more transistors to their designs. One year later Intel **orrered** the 1103 with 1024 **pits** of memory, and **ths rapioly** became a standard component in the **coraputerg** of the day. In April 1970, Intel hired **Raggin** the inventor of the silicon gate MOSFET, away from **Fairhifd**. On **Faggint** second day at Intel, **Magatosh** Shima, the engineering representative from **Bnsicom**, arrived **trom** Japan to review the **desigin**. |

**Table 5:** Results after Applying the Proposed Algorithm

|  |
|---|
| In 1968, Bob Noyce and Gordon Moore **leit** Fairchild Semiconductor **ro** start their own company focused on building products from integrated circuits. They named their company **Lintelo** (from INTegrated Electronics). In 1969, Intel began shipping the first commercial integrated circuit using MOSFETs, a 256-bit memory chip called the **L101**. The 1101 memory chip did not sell well, but Intel was able to rapidly shrink the size of the new silicon gate MOSFETs and add more transistors to their designs. One year later Intel offered the 1103 with 1024 bits of memory, and this rapidly became a standard component in the computers of the day. In April 1970, Intel hired **Raggin**, the inventor of the silicon gate MOSFET, away from Fairchild. On **Faggint** second day at Intel, Masatoslu Shima, the engineering representative from Busicom, arrived from Japan to review the design. |

The OCR text delineated in table 4 includes 30 OCR error words out of 141 total words (number of words in the whole text), making the error rate close to $E = 30/141 = 0.212 = 21.2\%$. Several of these errors are proper names such as "Noyce", "Moore", "Intel", and "Busicom"; whereas others are technical words such as "MOSFET" and "bits". The remaining errors are regular English words such as "building", "memory", "offered", "design", etc. Table 5 exposes the results of applying the proposed OCR error correction algorithm; it obviously managed to correct 24 misspelled words out of 30, leaving only 6 non-corrected errors: "Lntelo" was falsely corrected as "Lintelo", and "leit", "ro", "L101", "Raggin", and "Faggint" were not corrected at all. As a result, the error rate using the proposed algorithm is close to $E = 6/141 = 0.042 = 4.2\%$. Consequently, the improvement can be calculated as $I = 0.212/0.042 = 5.04 = 504\%$, that is increasing the rate of error detection and correction by a factor of 5.

Furthermore, another experiment was conducted using the OmniPage program on the French document which is depicted in Figure 3. It produced several OCR error words that are listed and underlined in Table 6. Table 7 shows the results of applying the proposed OCR error correction algorithm to detect and correct the misspellings present in table 6.

**Figure 3:**    Low-Quality French Image Document

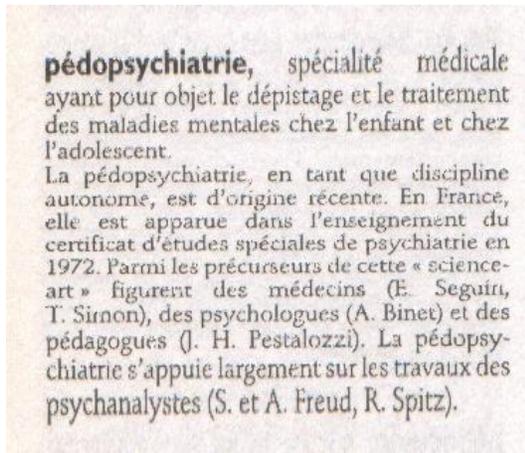

**Table 6:**    Results of Performing OCR using OmniPage

**pédopsychlatrie**, spécialité médicale ayant pour objet le dépistage **eu** le traitement des maladies mentales chez **P'enfant** et chez **l'aclolescent**. La pédopsychiatrie, en **tatlt** que discipline autonome, est d'origine récente. En France, elle est apparue dans **Penseignement** du certificat d'études spéciales de psychiatrie en 1972. Parmi les précurseurs de cette « science-art » **fignrent** des médecins (E **Seguini**, T. **Sirmon**), des psychologues (A. **Bineit**) et des pédagogues J. H. **Pestalozzij**. La pédopsy-chiatrie s'appuie largement sur les travaux des psychanalystes (S. et A. Freud, R. **Sp1tz**).

**Table 7:**    Results after Applying the Proposed Algorithm

pédopsychiatrie, spécialité médicale ayant pour objet le dépistage **eu** le traitement des maladies mentales chez l'enfant et chez l'adolescent. La pédopsychiatrie, en **tatli** que discipline autonome, est d'origine récente. En France, elle est apparue dans l'enseignement du certificat d'études spéciales de psychiatrie en 1972. Parmi les précurseurs de cette « science-art » figurent des médecins (E Seguin, T. Simon), des psychologues (A. Binet) et des pédagogues J. H. Pestalozzi. La pédopsy-chiatrie s'appuie largement sur les travaux des psychanalystes (S. et A. Freud, R. **Sp1tz**).

The OCR text delineated in table 6 includes 12 OCR error words out of 84 total words (number of words in the whole text), making the error rate close to $E = 12/84 = 0.142 = 14.2\%$. Several of these errors are proper names such as "Seguin", "Simon", "Binet", and "Pestalozzi"; whereas others are technical words such as "pédopsychiatrie". The remaining errors are regular French words such as "l'enfant", "l'adolescent", "figurent", etc. Table 7 exposes the results of applying the proposed OCR error correction algorithm; it obviously managed to correct 9 misspelled words out of 12, leaving only 3 non-corrected errors: "tatlt" was corrected as "tatli", and "eu" and "Sp1tz" were not corrected at all. As a result, the error rate using the proposed algorithm is close to $E = 3/84 = 0.035 = 3.5\%$. Consequently, the improvement can be calculated as $I = 0.142/0.035 = 4.05 = 405\%$, that is increasing the rate of error detection and correction by a factor of 4.

## 6. Conclusions
Experiments carried out on the proposed algorithm showed a drastic improvement in the detection and correction of OCR errors. Apparently, the proposed solution was able to detect and correct about 5 times (504%) more English errors and around 4 times (405%) more French errors than the OmniPage tool, yielding to an error rate close to 4.2% for the English text and 3.5% for the French text, compared with 21.2% for the English text and 14.2% for the French text using the OmniPage. Figure 4 and Figure 5 graphically represent the results of the experiments.

The foremost reason behind these remarkable results is the integration of Google Web 1T data set into the proposed algorithm as it contains wide-ranging, extensive, and up-to-date set of words and precise statistics about terms associations that cover domain specific words, technical terminologies, acronyms, expressions, proper names, and almost every word in the language.

**Figure 4:** Histogram Showing the Number of OCR Errors

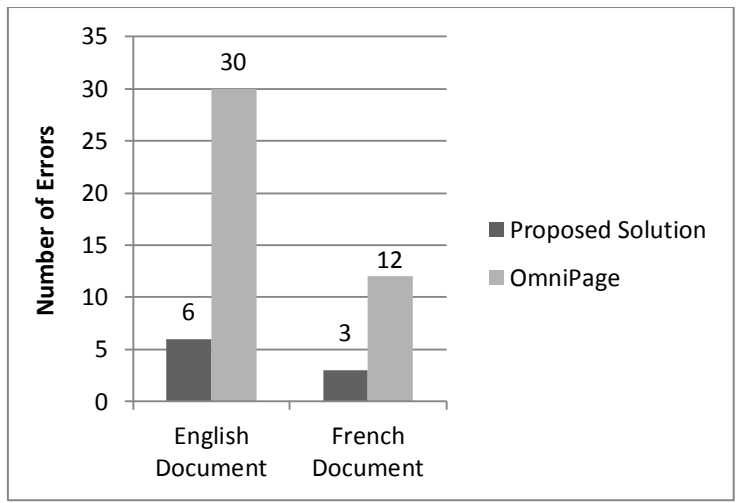

**Figure 5:** Histogram showing the OCR Error Rate

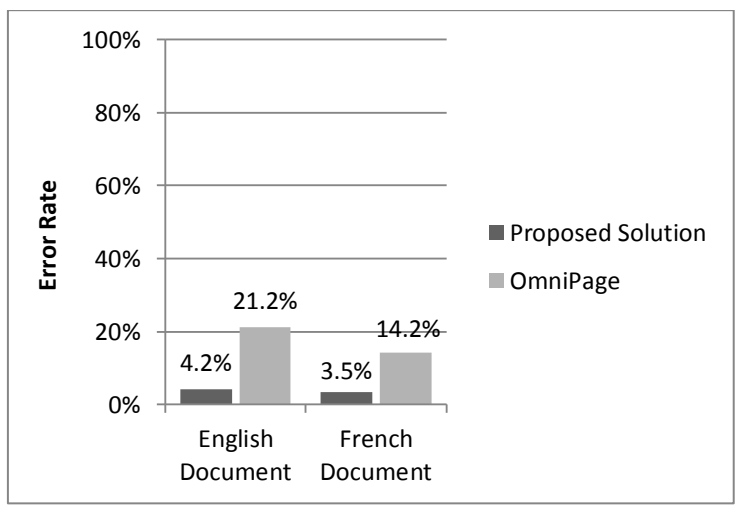

## 7. Future Work
As further research, other languages such as German, Arabic, and Japanese are to be experimented so as to internationalize the algorithm and extend its capability to treat documents written in different languages. Moreover, since the proposed algorithm is computationally intensive, optimizing it for parallel and distributed systems would certainly speed up its execution and processing time.

## Acknowledgment

This research was funded by the Lebanese Association for Computational Sciences (LACSC), Beirut, Lebanon under the "Web-Scale OCR Research Project – WSORP2011".